\documentclass[11pt,a4paper]{article}
\usepackage[hyperref]{naaclhlt2018}
\usepackage{times}
\usepackage{latexsym}
\usepackage{amsmath}
\usepackage[scaled=.8]{beramono}
\usepackage[T1]{fontenc}
\usepackage[utf8x]{inputenc}
\usepackage{multirow}
\usepackage{xcolor}
\usepackage{tikz-dependency}
\usepackage{url}
\usepackage{multirow}
\usepackage{graphicx}
\usepackage{subcaption}
\usepackage{arydshln}

\DeclareMathOperator*{\argmin}{argmin}
\aclfinalcopy
\usepackage{graphicx}
\newcommand{\heart}{\ensuremath\heartsuit}

\newcommand{\citeposs}[2][]{\citeauthor{#2}'s (\citeyear[#1]{#2})}
\newcommand{\Citeposs}[2][]{\Citeauthor{#2}'s (\citeyear[#1]{#2})}

\title{Parsing Tweets into Universal Dependencies}

\author{Yijia Liu \\
  Harbin Institute of Technology \\
  {\tt yjliu@ir.hit.edu.cn} \\\And
  Yi Zhu \\
  University of Cambridge \\
  {\tt yz568@cam.ac.uk} \\\AND
  Wanxiang Che \quad Bing Qin\\
  Harbin Institute of Technology\\\And
  Nathan Schneider \\
  Georgetown University \\\And
  Noah A. Smith \\
  University of Washington
  }

\date{}

\begin{document}
\maketitle
\begin{abstract}
We study the problem of analyzing tweets with
Universal Dependencies \citep[UD;][]{NIVRE16.348}. We extend the UD guidelines to cover
special constructions in tweets that affect tokenization,
part-of-speech tagging, and labeled dependencies. Using the extended guidelines, we create
a new tweet treebank for English ({\sc Tweebank v2}) that is four times larger than the (unlabeled) {\sc Tweebank
  v1} introduced by \citet{kong-EtAl:2014:EMNLP2014}. 
We characterize the disagreements between our annotators
and show that it is challenging to deliver
consistent annotation due to ambiguity in
understanding and explaining tweets. Nonetheless, using the new treebank,
we build a pipeline system to parse raw tweets into UD. To overcome 
annotation noise without sacrificing computational efficiency, we propose a new
method to distill an ensemble of 20 transition-based parsers into a single one. Our
parser achieves an improvement of 2.2 in LAS over the un-ensembled baseline 
and outperforms parsers that are state-of-the-art on other treebanks in both accuracy and speed.
\end{abstract}

\section{Introduction}
NLP for social media messages is challenging, requiring domain
adaptation and annotated datasets
(e.g., treebanks)
for training and evaluation.
Pioneering work by \citet{Foster:2011:HPT:2908630.2908634} 
annotated 7,630 tokens' worth of tweets according to the
phrase-structure conventions of the Penn Treebank
\citep[PTB;][]{Marcus93buildinga}, enabling conversion to Stanford Dependencies.
\citet{kong-EtAl:2014:EMNLP2014} further studied the challenges in
annotating tweets and presented a tweet treebank ({\sc Tweebank}),
consisting of 12,149 tokens and largely following conventions
suggested by \citet{schneider-EtAl:2013:LAW7-ID}, fairly close to 
\citet{Yamada03statisticaldependency} dependencies (without labels). 
Both annotation efforts were highly influenced by the PTB, whose guidelines
have good grammatical coverage on newswire. However, when it comes
to informal, unedited, user-generated text, the guidelines may leave
many annotation decisions unspecified.


Universal Dependencies \citep[UD]{NIVRE16.348} were introduced to enable
consistent annotation across different languages. To allow such
consistency, UD was designed to be adaptable to different genres \cite{wang-EtAl:2017:Long6} 
and languages \cite{guo-EtAl:2015:ACL-IJCNLP2,TACL892}. We propose that analyzing
the syntax of tweets can benefit from such adaptability. In this paper,
we introduce a new English tweet treebank of 55,607 tokens that follows the UD
guidelines, but also contends with social media-specific challenges that were not
covered by UD guidelines.\footnote{We developed our treebank independently of a similar effort for Italian tweets \cite{sanguinetti-17}. 
See \S\ref{sec:postwita} for a comparison.} Our annotation includes 
tokenization, part-of-speech (POS) tags, and (labeled) Universal Dependencies.
We characterize the disagreements among our annotators and find that
consistent annotation is still challenging to deliver even with
the extended guidelines.


Based on these annotations, we nonetheless designed a pipeline to parse 
raw tweets into Universal Dependencies. Our pipeline includes: a
bidirectional LSTM (bi-LSTM) tokenizer, a word cluster--enhanced POS
tagger \citep[following][]{owoputi-EtAl:2013:NAACL-HLT}, and a stack LSTM parser
with character-based word representations
\cite{ballesteros-dyer-smith:2015:EMNLP}, which we refer to as our
``baseline'' parser.
To overcome the noise in our annotated data and achieve better performance
without sacrificing computational efficiency, we 
distill a 20-parser ensemble into a single greedy  parser 
\cite{DBLP:journals/corr/HintonVD15}.
We show further  that learning directly from the exploration of the ensemble parser
is more beneficial than learning from the gold standard ``oracle''
transition sequence. Experimental results show that an improvement of more
than 2.2 points in LAS over the
baseline parser can be achieved with our distillation method.  It outperforms
other state-of-the-art parsers in both accuracy and speed.

The contributions of this paper include:
\begin{itemize}
\item We study the challenges of annotating tweets in UD (\S\ref{sec:anno})
and create a new tweet treebank ({\sc Tweebank v2}), which includes 
tokenization, part-of-speech tagging, and labeled Universal Dependencies.
We also characterize the difficulties of creating such annotation.

\item We introduce and evaluate a pipeline system to parse the raw tweet text into
Universal Dependencies (\S\ref{sec:parsing}).  Experimental results show
that it performs better than a pipeline of the state-of-the-art alternatives.

\item We propose a new distillation
method for training a greedy parser, leading to better performance
than existing methods and without efficiency sacrifices.
\end{itemize}

Our dataset and system are publicly available at \url{https://github.com/Oneplus/Tweebank}
and \url{https://github.com/Oneplus/twpipe}.

\section{Annotation}\label{sec:anno}

We first review {\sc Tweebank v1} of \citet{kong-EtAl:2014:EMNLP2014},
the previous largest Twitter dependency annotation effort 
(\S\ref{sec:tweebank}).
Then we introduce the differences in our tokenization
(\S\ref{sec:tok-anno}) and part-of-speech (\S\ref{sec:pos-anno}) (re)annotation with \citet{ICWSM101540} and 
\citet{gimpel-EtAl:2011:ACL-HLT2011}, respectively, on which {\sc Tweebank v1} was built. 
We describe our effort of adapting the
UD conventions to cover tweet-specific constructions (\S\ref{sec:ud-tweet}). 
Finally, we present our process of creating a new tweet treebank, 
{\sc Tweebank v2}, and characterize
the difficulties in reaching consistent annotations (\S\ref{sec:anno-process}).

\subsection{Background: \textsc{Tweebank}}\label{sec:tweebank}

The annotation effort we describe stands in contrast to the previous work
by \newcite{kong-EtAl:2014:EMNLP2014}.  Their aim was the rapid
development of a dependency parser for tweets, and to that end they
contributed a new annotated corpus, \textsc{Tweebank}, consisting of
12,149 tokens.  Their annotations added unlabeled dependencies to a portion of the data
annotated with POS tags by 
\newcite{gimpel-EtAl:2011:ACL-HLT2011} and
\newcite{owoputi-EtAl:2013:NAACL-HLT} after rule-based tokenization \citep{ICWSM101540}.
Kong et al.~also contributed a system for parsing;
we defer the discussion of their parser to \S\ref{sec:parsing}.

Kong et al.'s rapid, small-scale annotation effort was heavily constrained.  It was
carried out by annotators with only cursory training, no clear
annotation guidelines, and no effort to achieve consensus on controversial
cases. Annotators were allowed to underspecify their analyses.
Most of the work was done in a very short amount of
time (a day).  Driven both by the style of the text they sought to annotate
and by exigency, some of their annotation conventions included:
\begin{itemize}
\item Allowing an annotator to exclude tokens from the dependency
  tree.  A clear criterion for exclusion was not given, but many
  tokens were excluded because they were deemed ``non-syntactic.''
\item Allowing an annotator to merge a multiword expression into a
  single node in the dependency tree, with no internal structure.
  Annotators were allowed to take the same step with noun phrases.
\item Allowing multiple roots, since a single tweet might contain more
  than one sentence.
\end{itemize}
These conventions were justified on the grounds of making the
annotation easier for non-experts, but they must be revisited in our
effort to apply UD to tweets.

\subsection{Tokenization}\label{sec:tok-anno}
Our tokenization strategy lies between the strategy of
\citet{ICWSM101540} and that of UD.
There is a tradeoff between preservation of original tweet content and respecting
the UD guidelines.

The regex-based tokenizer of \citet{ICWSM101540}---which was 
originally designed for an exploratory search interface called
TweetMotif, not for NLP---preserves most whitespace-delimited tokens, including 
hashtags, at-mentions, emoticons, and unicode glyphs. 
They also treat contractions and acronyms as whole tokens and do not split them.
UD
tokenization,\footnote{\url{http://universaldependencies.org/u/overview/tokenization.html}}
in order to better serve dependency annotation, treats each syntactic word as a token.
They therefore more aggressively split
clitics from contractions (e.g., {\it  gonna} is tokenized as {\it gon} and {\it na}; {\it its}
is tokenized as {\it it} and {\it s} when {\it s} is a copula).
But acronyms are not touched
in the UD tokenization guidelines. Thus, we follow the UD tokenization for contractions
and leave acronyms like {\em idc} (``I don't care'') as a single token. 

In the different direction of splitting tokens, UD guidelines also suggest to merge
{\it multi-token words} (e.g., {\it 20 000}) into one single token in some special
cases. We witnessed a small number of tweets that contain multi-token words
(e.g., {\it Y O}, and {\it R E T W E E T}) but didn't combine them for simplicity.
Such tokens only account for 0.07\% and we use the UD
  {\it goeswith} relation to resolve these cases in the dependency annotations.

\subsection{Part-of-Speech Annotation}\label{sec:pos-anno}

Before turning to UD annotations, we (re)annotated the data with 
POS tags, for consistency with other UD efforts,
which adopt the universal POS tagset.\footnote{A revised and extended version of \citet{PETROV12.274} with 17 tags.}
In some cases,  non-corresponding tag conflicts arose between the UD English Web Treebank 
treebank conventions \cite[UD\_English-EWT;][]{Marneffe2014UniversalSD}\footnote{\url{https://github.com/UniversalDependencies/UD_English-EWT}}
and the conventions of \newcite{gimpel-EtAl:2011:ACL-HLT2011}.  
In these cases, we always
conformed to UD, enabling consistency (e.g., when we exploit the
existing UD\_English-EWT treebank in our parser for tweets, \S\ref{sec:parsing}).  For example,  
the nominal URL in Figure \ref{fig:informal-toks} is tagged as {\it
  other} ({\tt X}) and {\tt +} is tagged as {\it symbol} ({\tt SYM})
rather than {\it conjunction} ({\tt CCONJ}).  

Tokens that do not have a syntactic function (see Figure \ref{fig:non-syn-toks}, discussed at greater
length in the next section) were usually annotated as \emph{other}
(\texttt{X}), except for emoticons, which are tagged as \emph{symbol}
(\texttt{SYM}), following UD\_English-EWT.



Tokens that abbreviate multiple words (such as \emph{idc}) are resolved to the POS of the syntactic head of the
expression, following UD conventions (in this example, the head \emph{care} is
a verb, so \emph{idc} is tagged as a verb).
When the token is not phrasal, we use the POS of the left-most
sub-phrase.  For example, \emph{mfw} (``my face when'') is tagged as a
noun (for \emph{face}).  


Compared to the effort of
\newcite{gimpel-EtAl:2011:ACL-HLT2011}, our approach simplifies some
matters.  For example, if a token is not considered syntactic by UD
conventions, it gets an \emph{other} (\texttt{X}) tag (Gimpel et
al.~had more extensive conventions).  Other phenomena, like
abbreviations, are more complicated for us, as discussed above;
Gimpel et al.~used a single part of speech for such expressions.


Another important difference follows from the difference in
tokenization.  As discussed in \S\ref{sec:tok-anno}, UD calls for more
aggressive tokenization than that of \citet{ICWSM101540} which opted
out of splitting contractions and possessives. As a consequence of adopting \citeposs{ICWSM101540} tokenization, 
Gimpel et al.~introduced new parts of speech for these cases
instead.\footnote{These tags only account for 2.7\% of tokens,
  leading to concerns about data sparseness in tagging and all
  downstream analyses.}  For us, these tokens must be split, but
universal parts of speech can be applied.

\subsection{Universal Dependencies Applied to Tweets} \label{sec:ud-tweet}

We adopt UD version 2 guidelines to annotate the syntax of tweets.
In applying UD annotation conventions to tweets, the choices of 
\newcite{kong-EtAl:2014:EMNLP2014} must be revisited.  We consider the
key questions that arose in our annotation effort, and how we resolved them.



\paragraph{Acronym abbreviations.}  
We follow \citet{kong-EtAl:2014:EMNLP2014} and annotate the
syntax of an acronym as a single word without normalization. Their syntactic functions
are decided according to their context. \citet{eisenstein:2013:NAACL-HLT} studied the necessity of normalization
in social media text and argued that such normalization is problematic.
Our solution to the syntax of abbreviations follows the spirit of his argument. Because abbreviations which clearly carry
syntactic functions only constitute 0.06\% of the tokens in our
dataset, we believe that normalization for acronyms is an unnecessarily complicated
step.

\paragraph{Non-syntactic tokens.}  
\begin{table}
\centering
\begin{tabular}{lrr}
\multicolumn{2}{r}{syntactic (\%)} & non-syntactic (\%)\\ \hline
emoticons & 0.25 & 0.95 \\
RT & 0.14 & 2.49\\
hashtag & 1.02 & 1.24 \\
URL & 0.67 & 2.38 \\
truncated words & 0.00 & 0.49 \\
 \hline
 total & 2.08 & 7.55 \\
\end{tabular}
\caption{Proportions of non-syntactic tokens in our annotation. 
These statistics are obtained on 140 character--limited tweets. \label{tbl:non-synt-prop}}
\end{table}

\begin{figure*}[t]
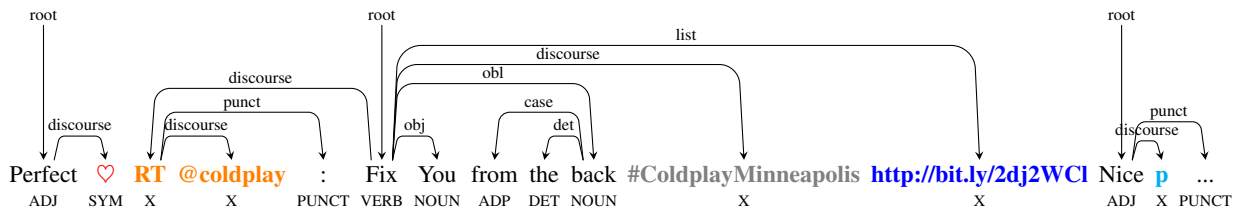

	\centering
	\small
	\begin{dependency}[edge slant=2, text only label, label style=above]
		\begin{deptext}
			Perfect \& \textcolor{red}{\bf $\heart$} \& \textcolor{orange}{\bf RT} \& \textcolor{orange}{\bf @coldplay} \& : \& Fix \& You \& from \& the \& back \& \textcolor{gray}{\bf \#ColdplayMinneapolis} \& \textcolor{blue}{\bf http://bit.ly/2dj2WCl} \& Nice \& \textcolor{cyan}{\bf p} \& ...\\
			\tiny ADJ \&\tiny SYM \&\tiny X \&\tiny X \&\tiny PUNCT \&\tiny VERB \&\tiny NOUN \&\tiny ADP \&\tiny DET \&\tiny NOUN \&\tiny X \&\tiny X \& \tiny ADJ \& \tiny X \& \tiny PUNCT \\
		\end{deptext}
		\deproot{1}{root}
		\depedge[edge unit distance=1em]{1}{2}{discourse}
		\deproot{6}{root}
		\depedge[edge unit distance=1em]{6}{3}{discourse}
		\depedge[edge unit distance=1em]{3}{4}{discourse}
		\depedge[edge unit distance=1em]{3}{5}{punct}
		\depedge[edge unit distance=1em]{6}{7}{obj}
		\depedge[edge unit distance=0.8em]{6}{10}{obl}
		\depedge[edge unit distance=1em]{10}{8}{case}
		\depedge[edge unit distance=1em]{10}{9}{det}
		\depedge[edge unit distance=0.8em]{6}{11}{discourse}
		\depedge[edge unit distance=0.8em]{6}{12}{list}
		\deproot{13}{root}		
		\depedge[edge unit distance=0.8em]{13}{14}{discourse}
		\depedge[edge unit distance=0.8em]{13}{15}{punct}
	\end{dependency}
	\caption{An example to illustrate non-syntactic tokens:
		\textcolor{red}{sentiment emoticon},
		\textcolor{orange}{retweet marker and its following at-mention},
		\textcolor{gray}{topical hashtag},
		\textcolor{blue}{referential URL}, and
		\textcolor{cyan}{truncated word}.
		This is a concatenation of three real tweets.
	}\label{fig:non-syn-toks}
\end{figure*}

\begin{figure*}[t]
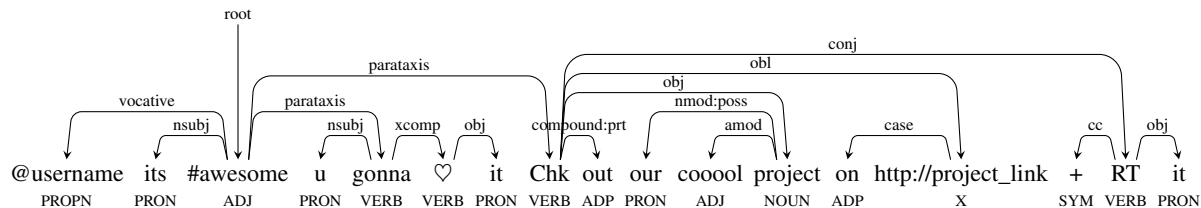

	\centering
	\small
	\begin{dependency}[edge slant=2, text only label, label style=above]
		\begin{deptext}
			@username \& its \& \#awesome \& u \& gonna \& $\heart$ \& it \& Chk \& out \& our \& cooool \& project \& on \& http://project\_link \& + \& RT \& it\\
			\tiny PROPN \& \tiny PRON \& \tiny ADJ \& \tiny PRON \& \tiny VERB \& \tiny VERB \& \tiny PRON \& \tiny VERB \& \tiny ADP \& \tiny PRON \& \tiny ADJ \& \tiny NOUN \& \tiny ADP \& \tiny X \& \tiny SYM \& \tiny VERB\& \tiny PRON\\
		\end{deptext}
		\deproot{3}{root}
		\depedge[edge unit distance=1em]{3}{1}{vocative}
		\depedge[edge unit distance=1em]{3}{2}{nsubj}
		\depedge[edge unit distance=1em]{3}{5}{parataxis}
		\depedge[edge unit distance=1em]{5}{4}{nsubj}
		\depedge[edge unit distance=1em]{5}{6}{xcomp}
		\depedge[edge unit distance=1em]{6}{7}{obj}
		\depedge[edge unit distance=0.7em]{3}{8}{parataxis}
		\depedge[edge unit distance=1em]{8}{9}{compound:prt}
		\depedge[edge unit distance=0.7em]{8}{12}{obj}
		\depedge[edge unit distance=1em]{12}{10}{nmod:poss}
		\depedge[edge unit distance=1em]{12}{11}{amod}
		\depedge[edge unit distance=0.6em]{8}{14}{obl}
		\depedge[edge unit distance=1em]{14}{13}{case}
		\depedge[edge unit distance=0.55em]{8}{16}{conj}
		\depedge[edge unit distance=1em]{16}{15}{cc}
		\depedge[edge unit distance=1em]{16}{17}{obj}
	\end{dependency}
	\caption{An example to illustrate informal but syntactic tokens.
		This is a contrived example inspired by several tweets.}\label{fig:informal-toks}
\end{figure*}

The major characteristic that distinguishes tweets from standard texts is that
a large proportion of tokens don't carry any syntactic function. 
In our annotation, there are five types of non-syntactic tokens commonly seen in tweets: 
sentiment emoticons, retweet markers and their following at-mentions, topical hashtags, referential URLs, and
truncated words.\footnote{The tweets we analyze have at most 140
  characters. Although Twitter has doubled the tweet length limit to
  280 characters since our analysis, we believe this type of token
  will still remain.}  Figure \ref{fig:non-syn-toks} illustrates examples of
these non-syntactic tokens.  As discussed above, these are generally
tagged with the \emph{other} (\texttt{X}) part of speech, except
emoticons, which are tagged as \emph{symbol} (\texttt{SYM}).  In our annotation, 
7.55\% of all tokens are belong to one of the five types; detailed statistics can be found in Table \ref{tbl:non-synt-prop}.

It is important to note that these types may, in some
contexts, have syntactic functions.
For example, besides being a discourse marker, \emph{RT} can
abbreviate the verb {\it retweet}; emoticons and hashtags may be
used as content words within a sentence; and at-mentions can be normal vocative proper nouns: see Figure~\ref{fig:informal-toks}.  
Therefore, the criteria for annotating a
token as non-syntactic must be context-dependent.

Inspired by the way UD deals with \emph{punctuation} (which is
canonically non-syntactic), we adopt the following
conventions
:

\begin{itemize}
\item If a non-syntactic token is within a sentence that has a clear predicate, it will be attached to this predicate. 
The retweet construction is a special case and we will discuss its treatment in the following paragraph.
\item If the whole sentence is a sequence of non-syntactic tokens, we attach all these tokens to the first one.
\item Non-syntactic tokens are mostly labeled as {\it discourse}, but
  URLs are always labeled as {\it list}, following the UD\_English-EWT dataset.
\end{itemize}

\newcite{kong-EtAl:2014:EMNLP2014} proposed an additional
preprocessing step, \emph{token selection}, in their annotation process.
They required the annotators to first select the non-syntactic
tokens and exclude them from the final dependency annotation.
In order to keep our annotation conventions in line with UD norms and preserve the original tweets as much as possible,
we include non-syntactic tokens in our annotation following the
conventions above. Compared with \citet{kong-EtAl:2014:EMNLP2014},
we also gave a clear 
definition of non-syntactic tokens, which helped
us avoid confusion during annotation.

\paragraph{Retweet construction.} Figure \ref{fig:non-syn-toks} shows
an example of the retweet construction (\emph{RT @coldplay :}).  This
might be treated as a verb phrase, with \emph{RT} as a verb and the
at-mention as an argument.
  This solution would lead to an uninformative root
word and, since this expression is idiomatic to Twitter, might create
unnecessary confusion for downstream applications aiming to identify
the main predicate(s) of a tweet.  We therefore treat the whole
expression as non-syntactic, including assigning the \emph{other}
(\texttt{X}) part of speech to both \emph{RT} and \emph{@coldplay},
attaching the at-mention to \emph{RT} with the \emph{discourse}
label and the colon to \emph{RT} with the \emph{punct}(uation) label,
and attaching \emph{RT} to the predicate of the following sentence.


\paragraph{Constructions handled by UD.}  A number of constructions
that are especially common in tweets are handled by UD
conventions: ellipsis, irregular word orders, and paratactic phrases
and sentences 
not explicitly delineated by punctuation.
	

\paragraph{Vocative at-mentions.}  Another idiomatic construction on
Twitter is a vocative at-mention (sometimes a
signal that a tweet is a reply to a tweet by the mentioned user). 
We
treat these at-mentions as vocative expressions, labeling them with
POS tag
\emph{proper noun} (\texttt{PROPN}) and attaching them to the
main predicate of the sentence it is within with the label \emph{vocative} as in UD guidelines (see Figure \ref{fig:informal-toks} for an example).

\subsection{Comparison to PoSTWITA-UD}\label{sec:postwita}
The first Twitter treebank annotated with Universal Dependencies 
was the \mbox{\textit{PosTWITA-UD}} corpus for Italian \citep{sanguinetti-17}, 
which consists of 6,738 tweets (119,726 tokens).
In their convention, tokenization tends to preserve the original
tweet content but two special cases, \textit{articulated prepositions}
(e.g., \textit{nella} as \textit{in la})
and \textit{clitic clusters} (e.g.~\textit{guardandosi} as \textit{guardando si}), are tokenized.
Their lemmas include spelling normalization, whereas our lemmas only normalize casing and inflectional morphology.
The current UD guidelines on lemmas are flexible, so variation between treebanks is expected.\footnote{\url{http://universaldependencies.org/u/overview/morphology.html\#lemmas}}

With respect to tweet-specific constructions,
\citeposs{sanguinetti-17} and our  interpretations of headedness are the same, 
but we differ in the relation label.
For topical hashtags, we use \textit{discourse} while they used \textit{parataxis}. 
In referential URLs, we use \textit{list} (following the precedent of UD\_English-EWT) while they used \textit{dep}.
Our choice of \textit{discourse} for sentiment emoticons
is inspired by the observation that emoticons are annotated as \textit{discourse} by UD\_English-EWT;
\citet{sanguinetti-17} used the same relation for the emoticons.
Retweet constructions and truncated words were not explicitly touched
by \citet{sanguinetti-17}.
Judging from the released treebank\footnote{\url{https://github.com/UniversalDependencies/UD_Italian-PoSTWITA}}, the \textit{RT} marker, at-mention, and colon in the retweet construction are all attached to the predicate of the following sentence with \textit{dep}, \textit{vocative:mention} and \textit{punct}.
We expect that the official UD guidelines will eventually adopt standards for these constructions so the treebanks can be harmonized.

\subsection{\sc Tweebank v2}\label{sec:anno-process}
Following the guidelines presented above, we create a new Twitter
dependency treebank, which we call {\sc Tweebank v2}.

\subsubsection{Data Collection}
{\sc Tweebank v2} is built on the original data of {\sc Tweebank v1}
(840 unique tweets, 639/201 for training/test), along with an 
additional 210 tweets sampled from the POS-tagged dataset of
\newcite{gimpel-EtAl:2011:ACL-HLT2011} and 2,500 tweets sampled
from the Twitter stream from February 2016 to July 2016.\footnote{Data downloaded from \url{https://archive.org/}.}
The latter data source consists of 147.4M English tweets after being
filtered by the {\it lang} attribute in the tweet JSON and {\it
  langid.py}.\footnote{\url{https://github.com/saffsd/langid.py}}
As done by \citet{kong-EtAl:2014:EMNLP2014}, the
annotation unit is always the tweet in its entirety---which may consist of multiple sentences---not the sentence alone. 
Before annotation, we use a simple regular expression to anonymize usernames and URLs.


\subsubsection{Annotation Process}
\begin{table}
	\centering
	\begin{tabular}{rrrrr}
		 & \multicolumn{2}{c}{\textsc{Tweebank v1}} & \multicolumn{2}{c}{\textsc{Tweebank v2}} \\
		split & tweets & tokens & tweets & tokens \\
		\hline
		train & 639 & 9,310 & 1,639 & 24,753 \\
		dev. & --\hphantom{0} & --\hphantom{00} & 710 & 11,742 \\
		test & 201 & 2,839 & 1,201 & 19,112 \\
		\hline
		total & 840 & 12,149 & 3,550 & 55,607 \\
	\end{tabular}

\caption{Statistics of {\sc Tweebank v2} and  comparison with
{\sc Tweebank v1}.
\label{tbl:data-stat}
}
\end{table}
Our annotation process was conducted in two stages.
In the first stage, 18 researchers worked on the {\sc Tweebank v1}
portion and the additional 210 tweets and created the initial annotations in one day.
Before annotating, they were given a tutorial overview of the general UD
annotation conventions and our guidelines specifically for annotating tweets.
Both the guidelines and annotations
were further refined by the authors of this paper to increase
the coverage of our guidelines and solve inconsistencies between
different annotators during this exercise. In the second stage, a tokenizer, a POS tagger, and a
parser were trained on the annotated data from the first stage (1,050 tweets in total),
and used to automatically analyze the sampled 2,500 tweets.  Authors 
of this paper manually corrected the parsed data and finally achieved 3,550 labeled tweets.\footnote{Manual annotation was done with Arborator \cite{gerdes:2013:W13-37}, a web platform for drawing dependency trees.}
Newly created annotations are split into train, development, and test sets and appended
to the original splits of {\sc Tweebank v1}. Statistics of our annotations and data splits are shown
in Table~\ref{tbl:data-stat}.


\begin{figure}
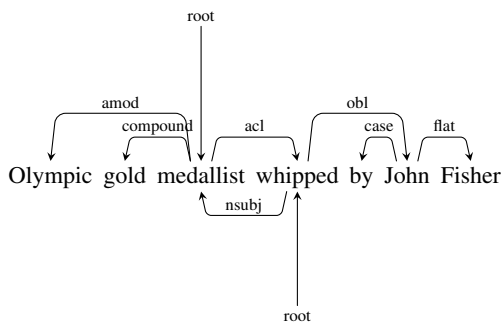

\centering
\small
\begin{dependency}[edge slant=2, text only label, label style=above]
	\begin{deptext}
		Olympic \& gold \& medallist \& whipped \& by \& John \& Fisher \\
	\end{deptext}
	\deproot{3}{root}
	\depedge[edge unit distance=1em]{3}{1}{amod}
    \depedge[edge unit distance=1em]{3}{2}{compound}
    \depedge[edge unit distance=1em]{3}{4}{acl}
    \depedge[edge unit distance=1em]{4}{6}{obl}
    \depedge[edge unit distance=1em]{6}{5}{case}
    \depedge[edge unit distance=1em]{6}{7}{flat}
	\deproot[edge below]{4}{root}
	\depedge[edge below, edge unit distance=1em]{4}{3}{nsubj}
\end{dependency}
\caption{An example of disagreement; one annotator's parse is shown above, disagreeing arcs from the other
  annotator are shown below. This is a real example in our annotation.}\label{fig:disagree}
\end{figure}

We report the inter-annotator agreement between the annotators
in the second stage. There is very little disagreement on the
tokenization annotation. 
The agreement rate is 96.6\% on POS, 
88.8\% on unlabeled dependencies, and
84.3\% on labeled dependencies. 
Further analysis shows the major disagreements on POS
involve entity names (30.6\%) 
and topical hashtags (18.1\%).
Taking the example in Figure \ref{fig:non-syn-toks}, ``Fix you'' 
can
be understood as a verbal phrase but also as the name of the Coldplay's
single and tagged as proper noun. 
An example of a disagreement on
dependencies is shown in Figure \ref{fig:disagree}.  Depending on
whether this is an example of a zero copula construction, or a clause-modified
noun, either annotation is plausible.

\section{Parsing Pipeline}
\label{sec:parsing}

We present a pipeline system to parse tweets into Universal
Dependencies.  We evaluate each component individually, and the system
as a whole.

\subsection{Tokenizer} \label{sec:tok}
Tokenization, as the initial step of many NLP tasks, is non-trivial for
informal tweets, which include hashtags, at-mentions, and emoticons
\cite{ICWSM101540}.  Context is often required for tokenization
decisions; for example,
the asterisk in {\it 4*3} is a separate token signifying
multiplication, but the asterisk in {\it sh*t}
works as a mask to evoke censorship and should not be segmented.

\begin{figure}[t]
	\centering
	\includegraphics[width=\columnwidth,trim={0 0 11cm 9cm},clip]{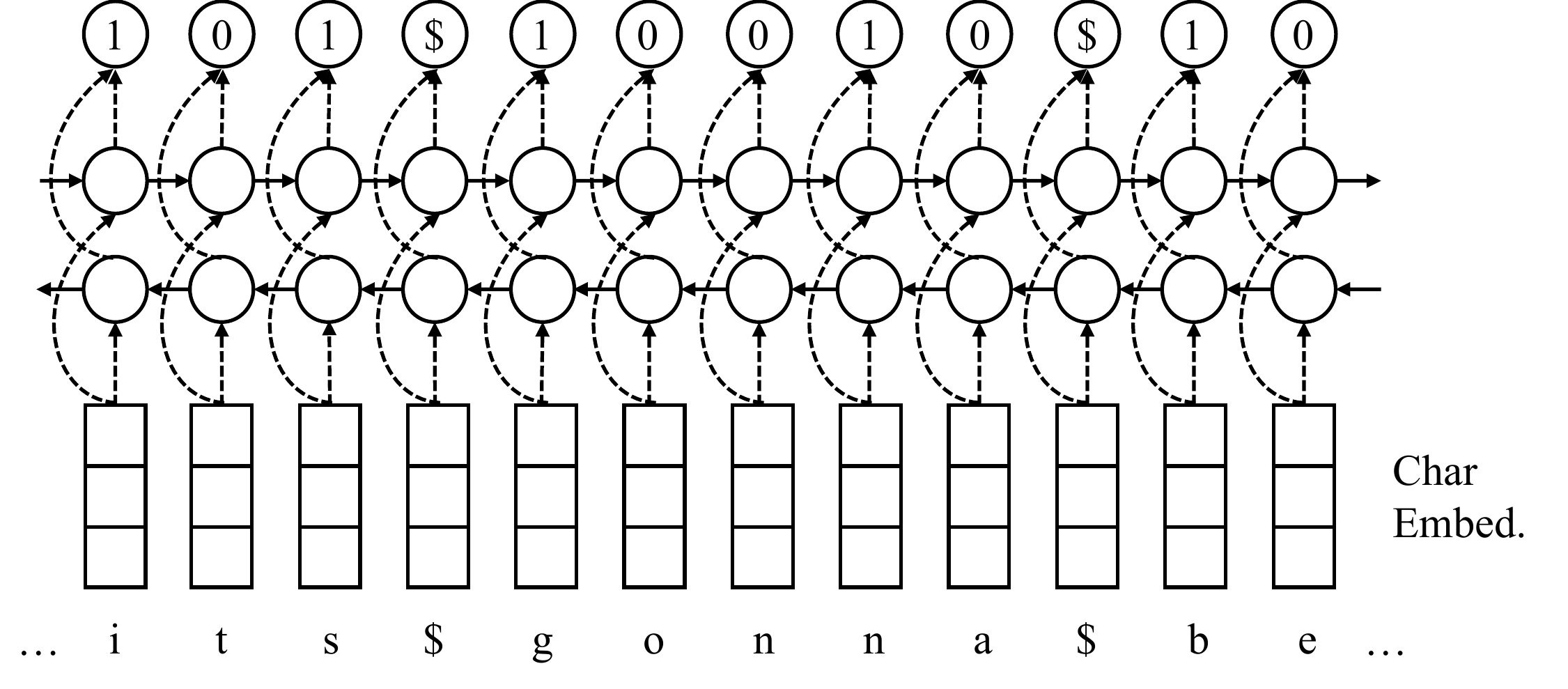}
	\caption{The bi-LSTM tokenizer that segments `{\it its gonna be}' into `{\it it s gon na be}'.}\label{fig:tok-model}
\end{figure}
We introduce a new
character-level bidirectional LSTM (bi-LSTM) sequence-labeling model
\cite{DBLP:journals/corr/HuangXY15,ma-hovy:2016:P16-1}
for tokenization.
Our model takes the raw sentence and tags each character in this 
sentence as whether it is the beginning of a word (1 as the beginning and 0 otherwise).
Figure \ref{fig:tok-model} shows the architecture of our tokenization model.
Space is treated as an input but deterministically  assigned a special tag \$.

\paragraph{Experimental results.}

Our preliminary results
showed that our model trained on
the combination of UD\_English-EWT and \textsc{Tweebank v2}
outperformed the one trained only on the UD\_English-EWT or \textsc{Tweebank v2},
consistent with previous work on dialect treebank parsing \cite{wang-EtAl:2017:Long6}.
So we trained our tokenizer on the training portion of
\textsc{Tweebank v2} combined with the UD\_English-EWT training set
and tested on the \textsc{Tweebank v2} test set. 
We report $F_1$ scores, combining precision and recall for token identification. Table \ref{tbl:tok-result} shows the
tokenization results, compared to  other available tokenizers. 
 Stanford CoreNLP \cite{manning-EtAl:2014:P14-5} and Twokenizer
\cite{ICWSM101540}\footnote{We use the updated version of Twokenizer from \citet{owoputi-EtAl:2013:NAACL-HLT}.} are rule-based systems and were not adapted
to the UD tokenization scheme.
The UDPipe v1.2
\cite{straka-strakova:2017:K17-3} model was re-trained on the same
data as our system. Compared with UDPipe, we use an LSTM
instead of a GRU in our model and we also use a larger size for hidden units (64 vs.~20),
which has stronger representational power. 
Our bi-LSTM tokenizer achieves the best accuracy among all these
tokenizers.  These results speak to the value of statistical modeling
in tokenization for informal texts.

\begin{table}[t]
	\centering
	\begin{tabular}{rc}
		\it System & $F_1$ \\
		\hline
		  Stanford CoreNLP & 97.3 \\
		 Twokenizer & 94.6 \\
		\hdashline
		UDPipe v1.2 & 97.4 \\
	 	our bi-LSTM tokenizer & 98.3 \\
	\end{tabular}
	\caption{Tokenizer comparison on the \textsc{Tweebank v2} test set.}\label{tbl:tok-result}
\end{table}

\subsection{Part-of-Speech Tagger}


Part-of-speech tagging for tweets has been extensively studied \cite{ritter-EtAl:2011:EMNLP,gimpel-EtAl:2011:ACL-HLT2011,derczynski-EtAl:2013:RANLP-2013,owoputi-EtAl:2013:NAACL-HLT,gui-EtAl:2017:EMNLP20172}. We therefore consider existing POS taggers for tweets instead of developing our own.
On the annotation scheme designed in \S\ref{sec:pos-anno}, based on UD and adapted for
Twitter, we compared several existing systems: the Stanford CoreNLP tagger, 
\citeposs{owoputi-EtAl:2013:NAACL-HLT} word cluster--enhanced tagger
(both greedy and CRF variants), and 
\citeposs{ma-hovy:2016:P16-1} neural network tagger 
which achieves the state-of-the-art performance on PTB.
\Citet{gui-EtAl:2017:EMNLP20172} presented a state-of-the-art neural tagger for Twitter,
but their implementation works only with the PTB tagset, so we exclude
it. 
All compared systems were re-trained on the combination of the UD\_English-EWT and 
\textsc{Tweebank v2} training sets. We use Twitter-specific GloVe embeddings released by
\citet{pennington-socher-manning:2014:EMNLP2014} in all neural taggers
and parsers.\footnote{\url{http://nlp.stanford.edu/data/glove.twitter.27B.zip}}

\paragraph{Experimental results.}

\begin{table}[t]
	\centering
	\begin{tabular}{rc}
		\it System & Accuracy \\
		\hline
		Stanford CoreNLP & 90.6 \\
		\citealp{owoputi-EtAl:2013:NAACL-HLT} (greedy) & 93.7 \\
		\citealp{owoputi-EtAl:2013:NAACL-HLT} (CRF) & 94.6 \\
		\hdashline
		\citealp{ma-hovy:2016:P16-1} & 92.5 \\
	\end{tabular}
	\caption{POS tagger comparison on gold-standard tokens in the
          \textsc{Tweebank v2} test set.
          \label{tbl:pos-result}}
\end{table}

\begin{table}[t]
	\centering
	\begin{tabular}{rc}
		\it Tokenization System & $F_1$ \\
		\hline
		Stanford CoreNLP & 92.3 \\
		our bi-LSTM tokenizer (\S\ref{sec:tok}) & 93.3 \\
	\end{tabular}
	\caption{\citet{owoputi-EtAl:2013:NAACL-HLT} POS tagging performance with automatic tokenization on
          the \textsc{Tweebank v2} test set. \label{tbl:pos-result-vs-tok}}
\end{table}

We tested the POS taggers on the \textsc{Tweebank v2} test set.  Results
with gold-standard tokenization are shown in
Table \ref{tbl:pos-result}. Careful feature engineering and 
\citet{Brown:1992:CNG:176313.176316} clusters  
help \citeposs{owoputi-EtAl:2013:NAACL-HLT} feature-based POS taggers to outperform \citeposs{ma-hovy:2016:P16-1} neural network
model. 

Results of the \citet{owoputi-EtAl:2013:NAACL-HLT}  tagger with non-greedy
inference on automatically tokenized data
are shown in  Table \ref{tbl:pos-result-vs-tok}.  We see that errors
in tokenization do propagate, but tagging performance is above 93\%
with our tokenizer. 

\subsection{Parser}

Social media applications typically require processing large volumes
of data, making speed an important consideration. We therefore 
begin with the neural greedy stack LSTM parser introduced by \citet{ballesteros-dyer-smith:2015:EMNLP},
which can parse a sentence in linear time and harnesses 
character representations to construct word vectors, which should help mitigate the challenge of
spelling variation. We encourage the reader to refer their paper for
more details about the model.

In our initial experiments, we train our parser on the combination of UD\_English-EWT
and \textsc{Tweebank v2} training sets. Gold-standard tokenization and automatic POS
tags are used. Automatic POS tags are assigned with 5-fold
jackknifing. Hyperparameters 
are tuned on the \textsc{Tweebank v2} development set. Unlabeled attachment score and
labeled attachment score (including punctuation) are reported.
All the experiments were run on a Xeon E5-2670 2.6 GHz machine.

\citet{reimers-gurevych:2017:EMNLP2017} and others have
pointed out that neural network training is 
nondeterministic and depends on the seed for the random
number generator.
Our preliminary experiments confirm this finding, with a gap of 1.4 LAS on development data
between the best (76.2)
 and worst (74.8) runs. To control for this
effect, we report the average of five differently-seeded runs, for
each of our models and the compared ones.

\begin{table}[t]
	\centering
	\begin{tabular}{rccc}
		\it System & UAS & LAS & Kt/s \\
		\hline
		\citet{kong-EtAl:2014:EMNLP2014} & 81.4 & 76.9 & 0.3 \\
		\citet{dozat-qi-manning:2017:K17-3} & 81.8 & 77.7 & 1.7 \\
		\hdashline
		\citet{ballesteros-dyer-smith:2015:EMNLP} & 80.2 & 75.7 & 2.3 \\
		Ensemble (20) & 83.4 & 79.4 & 0.2 \\
		Distillation ($\alpha =1.0$) & 81.8 & 77.6 & 2.3 \\
		Distillation ($\alpha =0.9$) & 82.0 & 77.8 & 2.3 \\		
		Distillation w/ exploration & 82.1 & 77.9 & 2.3 \\
	\end{tabular}
	\caption{Dependency parser comparison on \textsc{Tweebank v2} test set,
           with automatic POS tags. We use \citet{ballesteros-dyer-smith:2015:EMNLP}
           as our baseline and build the ensemble and distilling model over it.
           The ``Kt/s'' column shows
           the parsing speed evaluated by thousands of tokens
           the model processed per second.
            \label{tbl:parse-result}}
\end{table}

\paragraph{Initial results.}  The first section of Table~\ref{tbl:parse-result} compares
the stack LSTM with {\sc TweeboParser} (the system 
of \citealp{kong-EtAl:2014:EMNLP2014}) and the state-of-the-art parser
in the CoNLL 2017 evaluations, due to
 \citet{dozat-qi-manning:2017:K17-3}.
 \Citeposs{kong-EtAl:2014:EMNLP2014} parser is a graph-based
parser with lexical features and word cluster and it uses dual decomposition
for decoding. The parser in \citet{dozat-qi-manning:2017:K17-3} is also a graph-based parser
but includes character-based word representations and uses a biaffine classifier
to predict whether an attachment exists between two words.
Both of the compared systems require superlinear runtime due to graph-based parsing. 
They are re-trained on the same data as our
system.
Our baseline lags behind by nearly two LAS
points but runs faster than both of them.

\paragraph{Ensemble.}  Due to ambiguity in the training
data---which most loss functions are not robust to \citep{Frnay2014ClassificationIT}, including the log loss we use, following
\citet{ballesteros-dyer-smith:2015:EMNLP}---and due to the instability
of neural network training, we follow \citet{Dietterich2000} and
consider an ensemble of twenty parsers trained using different random
initialization.  To parse at test time, the transition probabilities of the twenty
members of the ensemble are averaged.  The result achieves LAS of
79.4, outperforming all three systems above (Table~\ref{tbl:parse-result}).

\paragraph{Distillation.}  The shortcoming of the 20-parser ensemble is, of
course, that it requires twenty times the runtime of a single greedy
parser, making it the slowest system in our comparison.  \newcite{kuncoro-16} proposed the distillation of 20
greedy transition-based parser into a single \emph{graph-based}
parser; they transformed the votes of the ensemble into a structured
loss function.  However, as Kuncoro et al.~pointed out, 
it is not straightforward to use a structured
loss in a \emph{transition-based} parsing algorithm.  Because fast
runtime is so important for NLP on social media, we
introduce a new way to distill our greedy ensemble into a single
transition-based parser (the first such attempt, to our knowledge).  

Our approach applies techniques from \newcite{DBLP:journals/corr/HintonVD15}
and \newcite{kim-rush:2016:EMNLP2016} to parsing.
Note that training a transition-based parser typically involves the
transformation of the training data into a sequence of ``oracle'' state-action
pairs.
Let $q(a \mid s)$ denote the distilled model's
probability of an action $a$ given parser state $s$; let $p(a\mid s)$ be the probability under the
ensemble (i.e., the average of the 20 separately-trained ensemble
members).
To train the distilled model, we minimize the interpolation between
their distillation loss and the conventional log loss:
\begin{align}
{\argmin}_q \quad  & \alpha \sum_i \underbrace{\sum_{a} -p(a
	\mid s_i) \cdot \log q(a \mid s_i)}_{\text{distillation loss}} \\
& +\ (1 - \alpha) \sum_i \underbrace{- \log q(a_i \mid
  s_i)}_{\text{log loss}} \nonumber
\end{align}

\begin{figure}[t]
	\centering
	\includegraphics[width=\columnwidth,trim={0.3cm 0 0 0},clip]{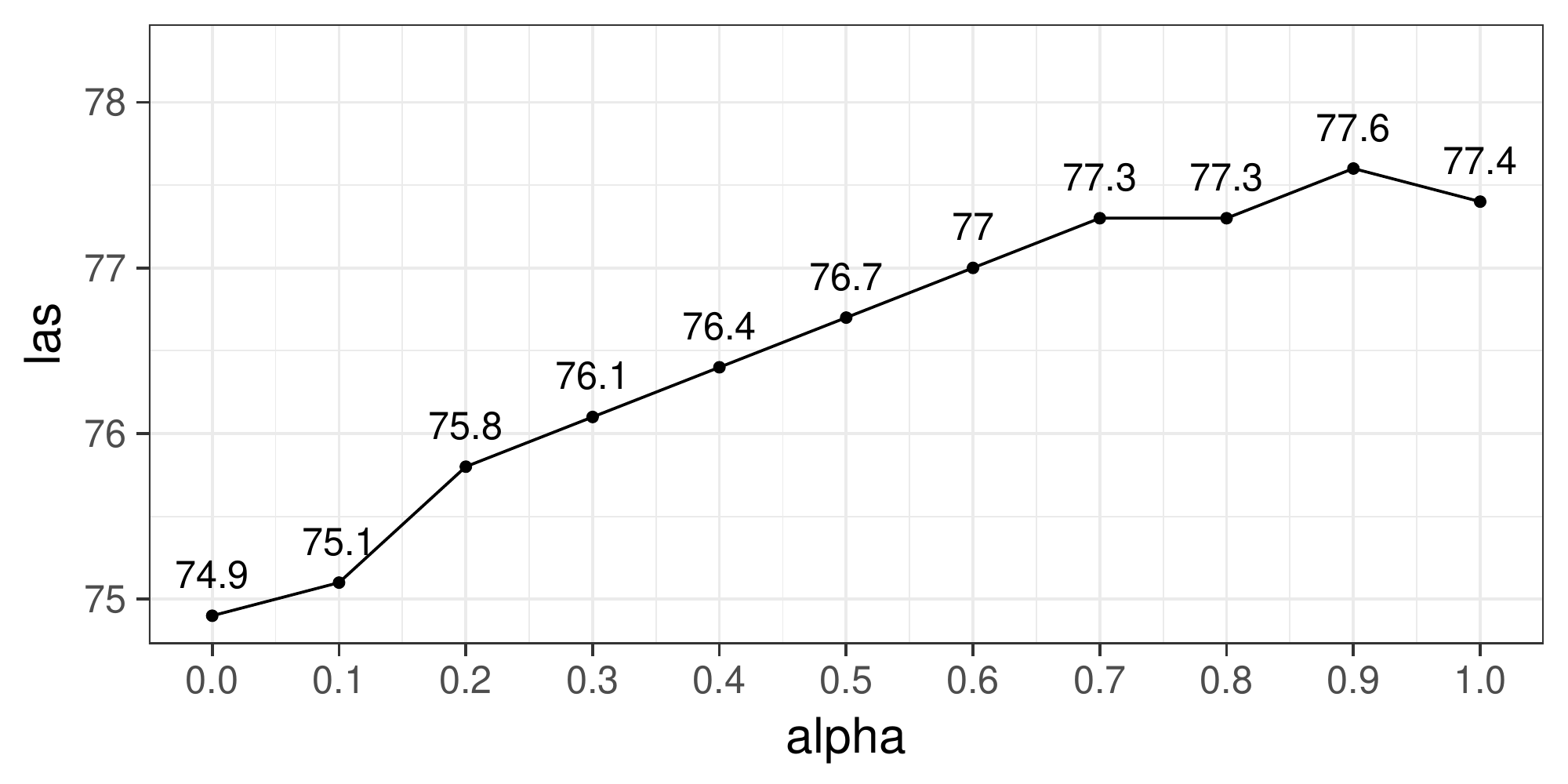}
	\caption{The effect of $\alpha$ on distillation.\label{fig:effect-alpha}}
\end{figure}

Distilling from this 
parser leads to a single greedy transition-based parser with 77.8
LAS---better than past systems but worse than our more expensive ensemble.
The effect of $\alpha$ is illustrated in
Figure~\ref{fig:effect-alpha}; generally paying closer attention to
the ensemble, rather than the conventional log loss objective, leads
to better performance.

\paragraph{Learning from exploration.} When we set $\alpha =1$, we
eliminate the oracle from the estimation procedure (for the distilled
model).  This presents an opportunity to learn with \emph{exploration}, by
randomly sampling transitions from the ensemble, found useful
in recent methods for training greedy models that use dynamic oracles
\citep{goldberg-nivre:2012:PAPERS,TACL145,TACL885,ballesteros-EtAl:2016:EMNLP2016}.  
We find that this
approach outperforms the  conventional distillation model, coming in
1.5 points behind the ensemble (last line of Table~\ref{tbl:parse-result}).

%


\begin{table}[t]
	\centering
	\begin{tabular}{rrcc}
		\it Pipeline stage & Score & Ours & SOTA \\
		\hline
		Tokenization  & \it $F_1$ & 98.3 & 97.3 \\		
		POS tagging &  \it  $F_1$ & 93.3 & 92.2 \\
		UD parsing & \it LAS $F_1$ & 74.0 & 71.4 \\
	\end{tabular}
	\caption{Evaluating our pipeline against a state-of-the-art pipeline.\label{tbl:pipline} 
	}
\end{table}

\paragraph{Pipeline evaluation.} Finally, we report  our full
pipeline's performance in  Table \ref{tbl:pipline}. We also compare
our model with a pipeline of the state-of-the-art systems (labeled  ``SOTA''):
Stanford CoreNLP tokenizer,\footnote{We choose the Stanford CoreNLP tokenizer in the spirit of comparing 
rule-based and statistical methods.} 
\citeposs{owoputi-EtAl:2013:NAACL-HLT} tagger, and \citeposs{dozat-qi-manning:2017:K17-3} parser.
Our system differs from the state-of-the-art pipeline in the tokenization and parser components.
From Table \ref{tbl:pipline}, our pipeline outperforms the state of
the art when
evaluated in pipeline manner.
The results also emphasize the
importance of tokenization:  without gold tokenization
UD parsing
performance drops by about four points.

\section{Conclusion}
We study the problem of parsing tweets into Universal Dependencies.
We adapt the UD guidelines to cover 
special constructions in tweets and create
the {\sc Tweebank v2}, which has 55,607 tokens. We characterize the disagreements
among our annotators and argue that inherent ambiguity in this genre
makes
consistent annotation a challenge.  Using this new treebank,
we build a pipeline system to parse tweets into UD. We also 
propose a new method to distill an ensemble of 20 greedy parsers into a single one
to overcome annotation noise without sacrificing efficiency.
Our parser achieves an improvement of 2.2  in LAS over a strong baseline
and outperforms other state-of-the-art parsers in both accuracy and speed.

\section*{Acknowledgments}
We thank Elizabeth Clark, Lucy Lin, Nelson Liu, Kelvin Luu, Phoebe Mulcaire, 
Hao Peng, Maarten Sap, Chenhao Tan, and Sam Thomson at the University of Washington, 
and Austin Blodgett, Lucia Donatelli, Joe Garman, Emma Manning, Angela Yang, and Yushi Zhang 
at Georgetown University for their annotation efforts in the first round.
We are grateful for the support from Lingpeng Kong at the initial stage of this project.
We also thank the anonymous reviewers for their helpful comments and suggestions.
This work was supported by the National Key Basic Research
Program of China via grant 2014CB340503 and the
National Natural Science Foundation of China (NSFC) via
grant 61632011.



\end{document}